\PassOptionsToPackage{table}{xcolor}
\documentclass[sigconf, nonacm]{acmart}

\AtBeginDocument{%
  }

\copyrightyear{2025}
\acmYear{2025}
\setcopyright{none}
\acmConference[DHOW '25]{Proceedings of the 2nd
International Workshop on Diffusion of Harmful Content on Online
Web}{October 27--28, 2025}{Dublin, Ireland}
\acmBooktitle{Proceedings of the 2nd International Workshop on Diffusion of
Harmful Content on Online Web (DHOW '25), October 27--28, 2025, Dublin,
Ireland}
\acmDOI{10.1145/3746275.3762209}
\acmISBN{979-8-4007-2057-4/2025/10}

\usepackage{booktabs}  
\usepackage{multirow}   
\usepackage{graphicx}
\usepackage{nicematrix}
\usepackage[table]{xcolor}
\definecolor{myred}{HTML}{cc4125}  
\definecolor{mygreen}{HTML}{6aa84f}  
\definecolor{myorange}{HTML}{fce5cd}  
\definecolor{mypurple}{HTML}{ff0000}  
\usepackage{balance}
\usepackage{tikz}
\usepackage{adjustbox}
\usepackage{makecell}
\usepackage{hyperref}

\begin{document}
\title{
Specializing General-purpose LLM Embeddings for Implicit Hate Speech Detection across Datasets}

\author{Vassiliy Cheremetiev}
\authornote{These authors contributed equally to this research.}
\orcid{0009-0003-3830-4536}
\affiliation{%
    \institution{EPFL}
    \city{Lausanne}
    \country{Switzerland}}
\affiliation{%
    \institution{Idiap Research Institute}
    \city{Martigny}
    \country{Switzerland}}
\email{v.cheremetiev@gmail.com}

\author{Quang Long Ho Ngo}
\authornotemark[1]
\orcid{0009-0009-2918-3385}
\affiliation{%
    \institution{EPFL}
    \city{Lausanne}
    \country{Switzerland}}
\affiliation{%
    \institution{Idiap Research Institute}
    \city{Martigny}
    \country{Switzerland}}
\email{quang.ngo@epfl.ch}

\author{Chau Ying Kot}
\orcid{0009-0009-2306-8722}
\affiliation{%
    \institution{EPFL}
    \city{Lausanne}
    \country{Switzerland}}
\email{chau-ying-kot@hotmail.com}

\author{Alina Elena Baia}
\orcid{0000-0001-5553-776X}
\affiliation{%
    \institution{Idiap Research Institute}
    \city{Martigny}
    \country{Switzerland}}
\email{alina.baia@idiap.ch}

\author{Andrea Cavallaro}
\orcid{0000-0001-5086-7858}
\affiliation{%
    \institution{EPFL}
    \city{Lausanne}
    \country{Switzerland}}
\affiliation{%
    \institution{Idiap Research Institute}
    \city{Martigny}
    \country{Switzerland}}
\email{andrea.cavallaro@epfl.ch}

\renewcommand{\shortauthors}{Vassiliy Cheremetiev, Quang Long Ho Ngo, Chau Ying Kot, Alina Elena Baia, \& Andrea Cavallaro}

\begin{abstract}
Implicit hate speech (IHS) is indirect language that conveys prejudice or hatred through subtle cues, sarcasm or coded terminology. IHS is challenging to detect as it does not include explicit derogatory or inflammatory words. To address this challenge, task-specific pipelines can be complemented with external knowledge or additional information such as context, emotions and sentiment data. In this paper, we show that, by solely fine-tuning recent general-purpose embedding models based on large language models (LLMs), such as Stella, Jasper, NV-Embed and E5, we achieve state-of-the-art performance. 
Experiments on multiple IHS datasets show up to 1.10 percentage points improvements for in-dataset, and up to 20.35 percentage points improvements in cross-dataset evaluation, in terms of F1-macro score.

\noindent \color{red} \textbf{Content warning}: This paper discusses examples of harmful text that may be offensive or upsetting.
\end{abstract}


\maketitle

\section{Introduction}
\label{sec:intro}
Hate speech detection is important to support content moderation in digital platforms, to foster inclusive discourse and to prevent social harm.

Hate speech can be explicit or implicit. Explicit hate speech (EHS) directly targets a protected entity and contains explicit keywords. Hence, early efforts for EHS detection primarily focused on identifying explicitly abusive language through keyword-based approaches~\cite{waseem-hovy-2016-hateful, davidson2017automated, schmidt-wiegand-2017-survey}.
IHS is "\textit{the use of coded or indirect language such as sarcasm, metaphor, and circumlocution to disparage a protected group or individual, or to convey prejudicial and harmful views about them}"~\cite{gao-etal-2017-recognizing, waseem-etal-2017-understanding, elsherief-etal-2021-latent}. 
IHS has a nuanced nature and manifests through a diverse range of subtle forms such as stereotypes, humor, and sarcasm~\cite{sap-etal-2020-social, davidson-etal-2019-racial, founta2018large, waseem-etal-2017-understanding, jurgens-etal-2019-just, qian-etal-2019-learning}.
Although IHS may not contain explicit hate words, it propagates prejudice and discrimination, and it is equally harmful as its explicit counterpart~\cite{basile-etal-2019-semeval, mozafari2020hate}. Even humans may struggle to understand the underlying meaning and intent behind such expressions~\cite{sap-etal-2020-social, hartvigsen-etal-2022-toxigen}.

Detecting IHS is made difficult by its lexical and semantic similarity to non-hateful content. IHS detection requires a nuanced understanding of implied meaning~\cite{masud2024focalinferentialinfusioncoupled}, real-world knowledge related to an event, specific social contexts, and the target.  

LLMs capture and represent extensive world knowledge~\cite{yu2023kola}, which could be leveraged for hate speech detection. 
Prior works explored prompting LLMs in scenarios like zero-shot~\cite{huang2023chatgpt, yang2023hare, hot_chatgpt, zhu2023chatgptreproducehumangeneratedlabels, damo}, zero-shot with chain-of-thought~\cite{yang2023hare}, and few-shot in-context learning~\cite{zhang2024dontextremesrevealingexcessive}. LLMs incorporate safeguards that prevent models from answering or discussing some sensitive topics like hateful content. Moreover, LLMs may exhibit limitations like excessive focus on sensitive groups, thus resulting in wrong classification of benign speech as \texttt{hate}, or extreme confidence score distributions resulting in poor calibration~\cite{zhang2024dontextremesrevealingexcessive}.
Overall, these models (e.g., GPT-3.5-Turbo, LLaMa2-7B, Mixtral-8x7b) typically underperform task-specific fine-tuned models~\cite{yang2023hare, zhang2024dontextremesrevealingexcessive, damo}.

In this paper, we evaluate fusing multiple sources of information to enhance BERT-based classifiers and leverage the ability of LLMs to generate contextual information for IHS detection. Specifically, we explore four fusion strategies to complement content information with contextual and emotion information. We find that while information fusion via feature concatenation provides a slight improvement over content-only BERT-based classifiers, fine-tuning general-purpose LLM-based embeddings (e.g.,~Stella~\cite{zhang2025jasperstelladistillationsota}, Jasper~\cite{zhang2025jasperstelladistillationsota}, NV-Embed~\cite{lee2025nvembedimprovedtechniquestraining},  E5~\cite{wang2024multilinguale5textembeddings}) allows us to reach new state-of-the-art performance for IHS detection. In summary, our main contributions are as follows: 
\begin{itemize}
    \item We present a comprehensive comparative evaluation of BERT-based and recent embedding-based classifiers, and show that fusion with LLM-generated context and emotion information can only marginally enhance the performance of a BERT-based classifier. We introduce new state-of-the-art benchmarks in this category of classifiers based on fine-tuning of generalist embedding models.
    \item We show that specializing embedding-based models significantly improves IHS detection in cross-dataset settings. This approach outperforms current state-of-the-art methods~\cite{kim2024labelawarehardnegativesampling, jiang-2025-learn, kim-etal-2023-conprompt, yang2023hare}  on several IHS datasets up to 1.10 percentage points for in-dataset evaluation and up to 20.35 percentage points for cross-dataset evaluation (F1-macro score). The significant improvement in cross-dataset evaluation is particularly noteworthy for generalization across datasets.
\end{itemize}    
Our approach is significant because it simplifies the detection process and eliminates the need for (explicit) external knowledge. To the best of our knowledge, we are the first to use general-purpose LLM-based embeddings models for IHS detection. The code is available at \href{https://github.com/idiap/implicit-hsd}{https://github.com/idiap/implicit-hsd}.

\section{Related Work}
\label{sec:related_work}

Early research in hate speech detection primarily focused on identifying explicit abusive language through linguistic features, such as character n-grams~\cite {waseem-hovy-2016-hateful} or word-centered features (i.e., literal words, part-of-speech tagging, occurrence of words within a word window)~\cite{warner-hirschberg-2012-detecting}. A combination of features such as TF-IDF weighted n-grams, part-of-speech tags, metadata including indicators for elements like hashtags and URLs, and number of characters and words was also used to train classifiers~\cite{davidson2017automated, schmidt-wiegand-2017-survey}. In~\cite{del2017hate}, the authors explore the combination of lexical and syntactic features with word sentiments and word embeddings.
These models rely on phrase structure and fail to capture the complexity and subtlety of the language used in social media. Transformer-based models have improved the quality of classification~\cite{mozafari2020bert, saleh2023detection}.
Later works~\cite{sap-etal-2020-social, davidson-etal-2019-racial, founta2018large, waseem-etal-2017-understanding, jurgens-etal-2019-just, qian-etal-2019-learning}  have emphasized the nuanced nature and complexity of implicit hate.  
Progress has been made in this area by focusing on specific types of implicit hate, such as euphemistic hate speech~\cite{magu-luo-2018-determining}, sarcasm detection~\cite{abu-farha-etal-2022-semeval}, as well as through multi-task learning~\cite{MIN2023214, 9509436, awal2021angrybertjointlearningtarget, mnassri2023hatespeechoffensivelanguage, jafari2023finegrained_emotions}, external knowledge integration~\cite{lin-2022-leveraging, sridhar-yang-2022-explaining, kim-etal-2023-conprompt, yang2023hare, perez2023assessing} or contrastive learning-based methods~\cite{ahn-etal-2024-sharedcon, kim2024labelawarehardnegativesampling, jiang-2025-learn, ocampo-etal-2023-unmasking}.

\noindent \textbf{Multi-task learning}. Classifiers can be trained to detect hate speech jointly with secondary tasks. For example, as hate speech may relate to emotions~\cite{fischer2018we}, a secondary task can be emotion classification~\cite{MIN2023214}. Plaza-Del-Arco et al.~\cite{9509436} achieves promising results on binary hate speech detection by combining sentiment and emotion into their features. 
Awal et al.~\cite{awal2021angrybertjointlearningtarget} 
employs a multitask learning approach to
jointly learn hate speech detection with secondary tasks, such as emotion classification and hateful target identification.  The authors use a BERT transformer~\cite{devlin2018bert} to share knowledge between tasks and Bidirectional Long-Short Term Memory Networks to learn task-specific representation, followed up by a gated fusion mechanism.
The authors base their approach on the intuition that datasets from relevant tasks can augment the hate speech data for the primary detection task. 
The method proposed in \cite{mnassri2023hatespeechoffensivelanguage} leverages emotion recognition as an auxiliary task for both hate speech and offensive language detection, via a shared BERT-based encoder and task-specific classification heads.
Similarly, Jafari et al.~\cite{jafari2023finegrained_emotions}  incorporates sentiment features alongside fine-grained emotion and textual features to improve the detection of IHS compared to single-task methods. 

\noindent \textbf{External knowledge}. 
Recent research focuses on enhancing hate speech detection by integrating various forms of real-world external knowledge (entity linking~\cite{lin-2022-leveraging}, knowledge bases~\cite{sridhar-yang-2022-explaining}).
Lin et al.~\cite{lin-2022-leveraging} 
links words appearing in tweets to their Wikipedia description and concatenates them with the original tweet before encoding.
Sridhar et al.~\cite{sridhar-yang-2022-explaining} combine explicit knowledge from knowledge bases with expert knowledge from high-quality annotation and LLM-generated knowledge to improve explanations of stereotypes in toxic speech.
Kim et al.~\cite{kim-etal-2022-generalizable} and Kim et al.~\cite{kim-etal-2023-conprompt} propose methods that utilize external knowledge, such as implications of anchor sentences and synonym substitution or machine-generated statements, respectively, to improve IHS detection using contrastive learning. 
In ~\cite{yang2023hare}, the authors incorporate explanations generated using chain-of-thought to better discern between \texttt{hate} and \texttt{not hate} and to improve generalization to unseen datasets.
Pérez et al. \cite{perez2023assessing} also demonstrates that hateful messages directed at certain communities, such as the LGBTI community, may benefit from the addition of context. The authors show that incorporating contextual parent comments and the corresponding news articles can improve the detection of hate speech in responses to posts from media outlets.

\noindent \textbf{Contrastive learning}.
Ahn et al. ~\cite{ahn-etal-2024-sharedcon} designed a clustering-based contrastive learning technique that uses shared semantics extracted from the data to learn discriminative representations. Specifically, the model is trained to pull together posts from the same cluster and push apart those from different clusters. This approach eliminates the need for costly human-annotated implications or machine-augmented data.
Kim et al.~\cite{kim2024labelawarehardnegativesampling} propose a contrastive learning-based approach that leverages hard negative samples to mitigate overfitting and improve generalization without relying on external knowledge. Building on this idea, Jiang et al.~\cite{jiang-2025-learn} use prediction errors to select hard positive samples
for contrastive learning to encourage the model to learn more robust representations 
to the spurious attributes that cause the misclassification. 

Ocampo et al.~\cite{ocampo-etal-2023-unmasking} use contrastive learning to bridge the representation gap between explicit and implicit hate speech. 
The authors build upon the observation that explicit and implicit text representations, when grouped by their target groups, tend to cluster together. 
The method pushes closer together pairs of implicit and explicit messages sharing the same target group, while pushing apart negative pairs (\texttt{hate} and \texttt{not hate} instances). 
This leads to more meaningful embedding representations and better separations between \texttt{not hate} and \texttt{hate} instances.
Masud et al.~\cite{masud2024focalinferentialinfusioncoupled} 
proposes to improve IHS detection by aligning the surface form of implicit hate with its implied meaning and increasing inter-cluster separation in the latent space to better distinguish speech categories.
%
\section{Models}
\subsection{Enhancing BERT-based classifiers}
\begin{figure}[t!]
    \centering
    \includegraphics[width=1\linewidth]{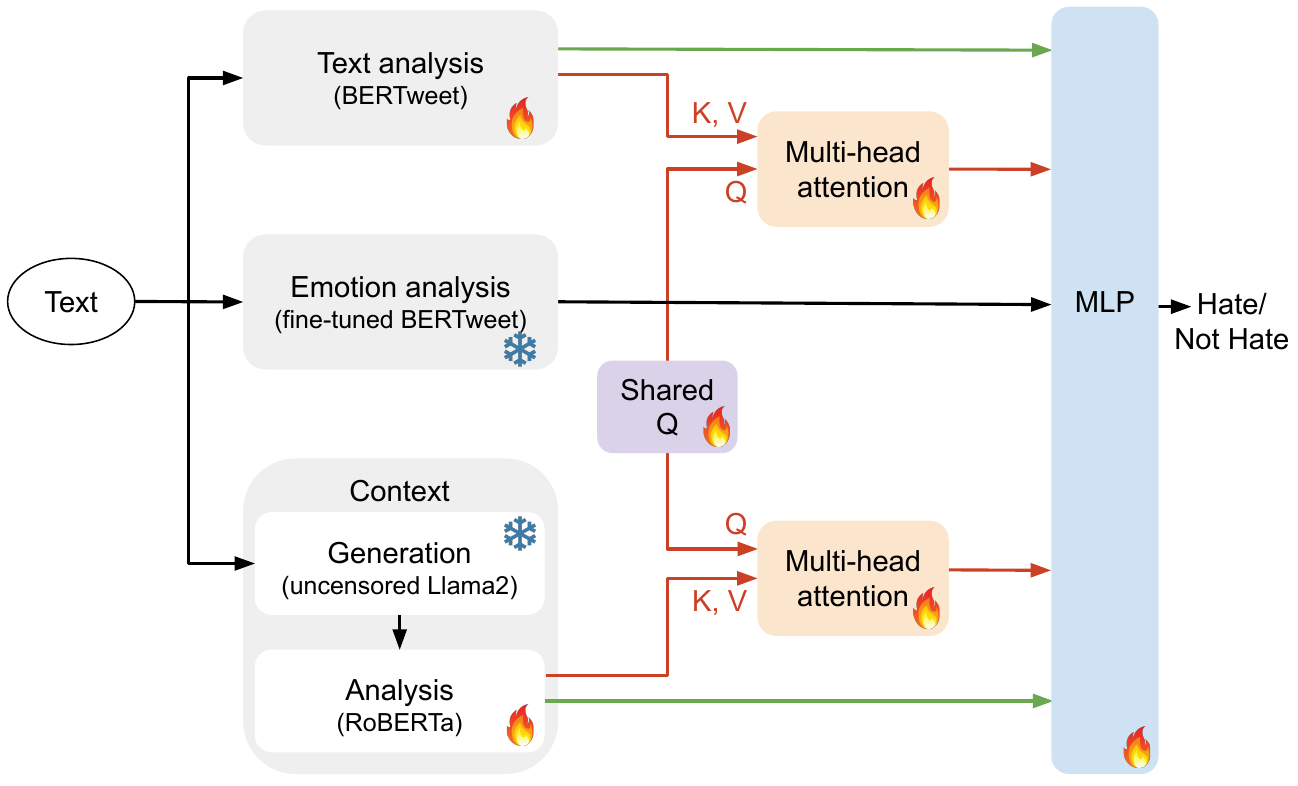}
    \caption{ Overview of our BERT models. 
    In gray, the three main components: tweet, emotion, and context module. In orange/purple, the added components for the shared learnable query architecture.
    \textcolor{mygreen}{Green}/\textcolor{myred}{red} arrows show the information flow for the \textcolor{mygreen}{concatenation}/\textcolor{myred}{shared query} fusion, respectively.
    Emotion features are directly fed to the MLP for both strategies. Q, K, V represent the query, key and value.
    }
    \Description{A diagram showing the workflow of the BERT-based models using feature fusion via concatenation and shared learnable query. }
    \label{fig:model_architecture}
\end{figure}

BERT~\cite{devlin2018bert} and its variants such as RoBERTa~\cite{liu2019roberta} have been extensively used for text classification~\cite{aragon-etal-2023-disorbert}. Hate speech detection works~\cite{kim-etal-2023-conprompt,jiang-2025-learn, ahn-etal-2024-sharedcon, kim2024labelawarehardnegativesampling, kim-etal-2022-generalizable} predominantly use models such as BERT, RoBERTa, and T5~\cite{T5}. 
Table~\ref{tab:backbones} shows a summary of the backbone architectures used by the most recent related works on IHS.
\begin{table}[t!]
\small

\caption{Backbone models used  for IHS detection. Multiple models indicate variations in the original work.}

\begin{tabular}{ll}
\toprule
{\bf Backbone model} & {\bf  Related work} \\ 
\midrule
BERT           & \begin{tabular}[c]{@{}l@{}}ImpCon~\cite{kim-etal-2022-generalizable}, LAHN~\cite{kim2024labelawarehardnegativesampling}, SharedCon~\cite{ahn-etal-2024-sharedcon},\\ CCL~\cite{jiang-2025-learn}, ConPrompt~\cite{kim-etal-2023-conprompt}, MTL~\cite{mnassri2023hatespeechoffensivelanguage},\\ AngryBERT~\cite{awal2021angrybertjointlearningtarget}, FiADD~\cite{masud2024focalinferentialinfusioncoupled}, EHSor~\cite{MIN2023214}\end{tabular} \\
& Contrastive BERT~\cite{ocampo-etal-2023-unmasking} \\

RoBERTa        & ImpCon~\cite{kim-etal-2022-generalizable}, LAHN~\cite{kim2024labelawarehardnegativesampling}  \\

HateBERT       & ImpCon~\cite{kim-etal-2022-generalizable}, CCL~\cite{jiang-2025-learn}, FiADD~\cite{masud2024focalinferentialinfusioncoupled},  \\

      &  Contrastive HateBERT~\cite{ocampo-etal-2023-unmasking}   \\

mBERT          & MTL~\cite{mnassri2023hatespeechoffensivelanguage}                         \\ \bottomrule
\end{tabular}
\label{tab:backbones}
\end{table}
We enhance the BERT model by incorporating tweet-level emotion information and tweet-driven contextual information via dedicated modules. Our BERT-based classifiers consists of three main components, namely text analysis, emotion analysis, and context generation (see Figure~\ref{fig:model_architecture}). 

\noindent \textbf{Feature extraction}.
The {\em text analysis} module uses a fine-tuned BERT to extract the content of the tweet and represent it into an embedding vector. 
The {\em emotion analysis} module infers with a fine-tuned BERTweet~\cite{pérez2023pysentimiento} a vector of probabilities across the following classes: fear, disgust, surprise, anger, sadness, joy, or other. Using a vector of probabilities instead of a single class allows the model to capture the complexity of the emotion.
Understanding IHS relies heavily on contextual nuances.  Capturing relevant context is made challenging by the short text length (tweets). Our {\em context} module leverages uncensored Llama2\footnote{\url{https://huggingface.co/georgesung/llama2_7b_chat_uncensored}} to generate the associated context, avoiding safeguards that might prevent processing and generation of certain content. 
We prompt the LLM to produce a neutral and factual context, which may include historical background or descriptions of stereotypes concerning the target of the text:

Prompt: \textit{As an educational assistant, your task is to provide neutral and objective analysis of the provided tweet, without any personal biases. 
        Offer short and concise information, context, and concepts to understand the content of the tweet without bias.
        The tweet may originate from different extremist groups, including White Nationalist, Neo-Nazi, Anti-Immigrant, Anti-Muslim, Anti-LGBTQ, KKK as well as non-extremist sources.
        The tweet could contain sarcasm, stereotypes, satire, metaphor, irony, or misinformation.
        Remember to avoid injecting personal opinions or interpretations into your analysis.
        Your aim is to provide a neutral understanding of the tweet's content within a maximum of 150 words. }
        
The final prompt is [Prompt. "Tweet to analyze: ", <Original tweet>.].
We explicitly ask for an objective and neutral analysis to try to avoid bias from the data Llama2 was trained on. We also give a context about the dataset that is used so the LLM has a starting point (see Appendix \ref{app:appendix_context generation} for examples of generated context).
The generated context is then used by RoBERTa to extract features.

\noindent \textbf{Feature fusion}. We explore four feature fusion approaches, namely concatenation, adaptive fusion, mixture of experts, and shared learnable query. 
With {\em concatenation}, we classify with a two-layer perceptron (MLP) the outputs of the three modules stringed together. The first layer of the MLP has the same size as the concatenated embeddings (1543), whereas the second layer contains 2 nodes for the binary classes.

With {\em adaptive fusion},  we learn the parameters $\alpha_\textit{tweet}$,  $\alpha_\textit{context}$, and $\alpha_\textit{emotion}$ that determine the scaling of each feature component. In order to maintain reasonable magnitude in the inputs, we constrain to $[-1,1]$ the learnable parameters with a sigmoid. 
With a simple {\em mixture of experts}, given a short text input, we utilize a simple MLP followed by a softmax layer to generate three adjustable feature scaling factors: $\alpha_\textit{tweet}$, $\alpha_\textit{context}$, $\alpha_\textit{emotion}$. The key distinction from adaptive fusion lies in the ability to tailor these scaling parameters specifically for each input, whereas adaptive fusion employs a fixed set of scaling parameters across all samples in the test dataset. Finally, for the {\em shared learnable query}, we use a multi-head attention with a shared learnable query, where keys and values are derived from both content and context embeddings. The query is a learnable parameter that is the same for both the content and context. The outputs of the multi-head attention blocks are then concatenated along with the emotion vector and fed to the classifier. 

\subsection{Specializing generalist embeddings}

General text embedding models, such as Stella~\cite{zhang2025jasperstelladistillationsota}, E5~\cite{wang2024multilinguale5textembeddings}, NV-Embed~\cite{lee2025nvembedimprovedtechniquestraining}, and Jasper~\cite{zhang2025jasperstelladistillationsota} are the result of numerous improvements over BERT \cite{devlin2018bert} and RoBERTa \cite{liu2019roberta}. Several factors contribute to the better performance of newer embedding models compared to BERT. First, the embedding models are trained on a bigger volume of data than BERT, enabling them to capture more diverse linguistic patterns and contextual nuances. Secondly, techniques such as hard-negative, in-batch negative and contrastive learning in general appear to provide better embeddings for classification even without a task specific pipeline for classification.
E5~\cite{wang2024multilinguale5textembeddings} is initialized from XLM-RoBERTa-large \cite{conneau-etal-2020-unsupervised} and results from curated datasets and contrastive learning with mined hard negatives.  
NV-Embed \cite{lee2025nvembedimprovedtechniquestraining}, a fine-tuned version of Mistral 7B \cite{jiang2023mistral7b}, is trained with contrastive learning using in-batch hard negatives and uses a latent attention layer to produce embeddings. 
Stella~\cite{zhang2025jasperstelladistillationsota} is based on mGTE \cite{zhang2024mgtegeneralizedlongcontexttext} and the general text embedding variant of Qwen2 \cite{yang2024qwen2technicalreport} where a final training involves matryoshka representation learning (MRL) \cite{kusupati2024matryoshkarepresentationlearning} which makes it performant at different embedding sizes. Jasper \cite{zhang2025jasperstelladistillationsota} uses a distillation of multiple teachers \cite{zhang2025jasperstelladistillationsota, lee2025nvembedimprovedtechniquestraining} and is augmented with multi-modal capabilities through a final training stage where image-caption pairs are used with SigLIP \cite{tschannen2025siglip2multilingualvisionlanguage} as the image encoder. These models also come in different sizes, with E5-large at 560 million parameters, Stella at 1.5 billion, Jasper at 2 billion, and NV-Embed at 7 billion.

\begin{figure}[t!]
    \centering
    \includegraphics[width=1\linewidth]{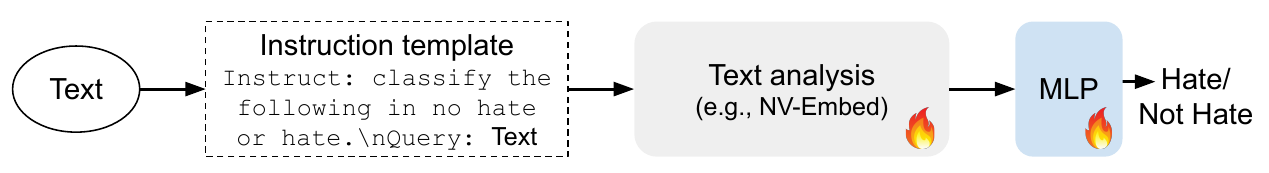}
    \caption{ Overview of the  embedding-based models. Given a task specific instruction, the generalist embeddings models are fine-tuned on the IHS datasets.  
    }
    \Description{A diagram showing the workflow of the generalist embeddings models. }
    \label{fig:model_architecture_gte}
\end{figure}

To remove instruction bias, all models are fine-tuned using the following instruction template: \texttt{Instruct: classify the following in no hate or hate.\textbackslash nQuery:}. The instruction is prepended to the short text that is being classified and then passed to the general text embedding model. Each model produces embeddings in $\mathbb{R}^{k \times n}$ whose dimensions depend on their specific implementation and the input length $k$. 
Following the recommendations provided by the model authors \footnote{\url{https://huggingface.co/NovaSearch/stella_en_1.5B_v5}} \footnote{\url{https://huggingface.co/intfloat/e5-large}}, we combine these embeddings into a single representation using a normalized sum over the token dimension. NV-Embed uses mean pooling as part of its final layer, we therefore use the output directly. This results in a final embedding vector in $\mathbb{R}^n$, which is subsequently fed into the classification module, which consists of a two-layer MLP with a hidden layer of size $n$ and LeakyReLU activations. The MLP ultimately reduces the dimensionality to 2 for classification (see Figure~\ref{fig:model_architecture_gte}).

To contrast the results of our embeddings-based classifiers, we compare them with linear probing (i.e., only the classification module is optimized)  and to recent generative models, such as  Llama3-8B \cite{grattafiori2024llama3herdmodels}, Gemma-7B \cite{gemmateam2024gemmaopenmodelsbased}, and Qwen3-8B \cite{yang2025qwen3technicalreport}.
For these LLMs, we take the average over the last hidden state as our embeddings \cite{wang2024textembeddingsweaklysupervisedcontrastive}  which are then fed to the same classification module as for the generalist embedding models. We fine-tune the whole pipeline.

\section{Validation} 
\label{sec:validation}

\subsection{Datasets} 

To quantify the performance of the classifiers, we employ four commonly used  IHS datasets that cover a variety of contexts and nuances of real-world scenarios. 
The distribution of labels in each dataset is reported in Table~\ref{tab:label_distribution_tab}.

\noindent\textbf{Implicit Hate Corpus} (IHC)~\cite{elsherief-etal-2021-latent}. This dataset consists of tweets collected between 2015 and 2017
from accounts of US extremist groups, including Black Separatist, White Nationalist, Neo Nazi, Anti-Muslim, Racist Skinhead, Ku Klux Klan, Anti-LGBT and Anti-Immigrant. Most of their speech targets minorities or specific groups of people.  The samples are labeled as \texttt{explicit hate}, \texttt{implicit hate}, or \texttt{not hate}. It is important to note the class imbalance in this dataset: 13206 tweets are \texttt{not hate} and 5460 contain \texttt{implicit hate}. 
Following~\cite{kim-etal-2022-generalizable, kim2024labelawarehardnegativesampling}, we only used the \texttt{implicit hate} samples in the dataset as the \texttt{hate} class, meaning that we do not use the \texttt{explicit hate} samples. 
An example of \texttt{not hate} sample is: \textit{"i have no idea what you are talking about. white supremacy = pure evil"}. An example of \texttt{implicit hate} sample is \textit{"\#hannahcornelius - why not come home to \#europe whites will never be welcome in \#southafrica"}.

\noindent\textbf{DynaHate} \cite{vidgen2021learning}. This dataset is built with an iterative process between a model and human annotators who progressively generate more challenging examples to trick the model (i.e.,~by flipping labels with minimal changes to the original post). The examples that are successful in tricking the model are then added to the training set. The model used for classification is RoBERTa with a sequence classification head, which is used to evaluate the difficulty of samples. The labeling includes \texttt{hate}/\texttt{not hate}, type of hate (e.g., threat, dehumanization), and target of \texttt{hate}. There are 41,255 entries, with $54\%$ of them labeled as \texttt{hate}. 

\noindent\textbf{SBIC}  \cite{sap-etal-2020-social}. This dataset contains social media posts from Reddit and Twitter with implicit social biases, stereotypes, and power dynamics in language. It was annotated by Amazon Mechanical Turk workers. The main labels contain: offensive/not offensive/maybe offensive, and secondary labels and annotations are: intend to offend, sexual content, group/individual targeting, targeted group, implied statement, in-group language (target of the same group as the writer). We follow \cite{kim-etal-2022-generalizable} and classify the text as \texttt{hate} if the aggregated score for offensiveness is equal to or above $0.5$.

\noindent\textbf{ToxiGen}~\cite{hartvigsen-etal-2022-toxigen}. This is a machine-generated dataset with toxic and benign statements about 13 minorities (e.g.,~African Americans, women, LGBTQ+). A subset of the generated data is validated by human annotators in terms of difficulty and toxicity. We use this subset, which is composed of 8960 training samples with 3368 being \texttt{hate}, 1792 validation samples with 638 \texttt{hate}, and 940 test samples among which 406 are \texttt{hate}. We use the split provided by the authors. We follow the indication from the official implementation\footnote{\url{https://github.com/microsoft/TOXIGEN}} and label a sample as \texttt{hate} if the sum of the toxicity score given by both the human and the model exceeds $5.5$.


\begin{table}[t]
  \small
  \centering
  \caption{Distribution of labels in the datasets.}
  \begin{NiceTabular}{lrrr}
    \toprule
    \textbf{Dataset} & \textbf{\# Samples} & \texttt{\textbf{Hate}} & \texttt{\textbf{Not hate}} \\
    \midrule
    IHC~\cite{elsherief-etal-2021-latent}      & 18666  & 5460  & 13206 \\
    Dynahate~\cite{vidgen2021learning}         & 41144  & 22175 & 18969 \\
    SBIC~\cite{sap-etal-2020-social}           & 44781  & 24048 & 20733 \\
    ToxiGen~\cite{hartvigsen-etal-2022-toxigen}&  9900  & 3774  &  6126 \\
    \bottomrule
  \end{NiceTabular}
  
  \label{tab:label_distribution_tab}
\end{table}

\subsection{Experimental setup}
\setlength{\tabcolsep}{4pt}
\begin{table*}[h!]
    \centering
    \small
    \caption{Results on IHC~\cite{elsherief-etal-2021-latent}, SBIC~\cite{sap-etal-2020-social}, Dynahate~\cite{vidgen2021learning} and ToxiGen~\cite{hartvigsen-etal-2022-toxigen} datasets for binary classification with \texttt{hate} as the positive class. We report the average over 5 runs with different seeds, the standard deviation for each metric is in parentheses. Models E5, Stella, Jasper and NV-Embed only use the tweet. Best result for each dataset/metric combination is in bold. Key- Acc: unweighted accuracy, P: precision, R: recall, F1-w: weighted F1-score, F1-M: macro F1-score, C: context features, E: emotion features, +: concatenation, AF: adaptive fusion, MoE: simple mixture of experts, SLQ: shared learnable query.}

    \begin{NiceTabular}{clccc|ccc|ccccc}
        \CodeBefore
            \cellcolor[gray]{0.9}{3-1,4-1,5-1,6-1, 7-1, 8-1, 9-1, 10-1, 11-1,12-1,13-1, 14-1, 15-1, 16-1, 17-1, 18-1, 19-1, 20-1,21-1,22-1, 23-1, 24-1, 25-1, 26-1}
        \Body
    \toprule
        & \textbf{Model} & \multicolumn{3}{c}{\texttt{\textbf{Not hate}}} & \multicolumn{3}{c}{\texttt{\textbf{Hate}}} & \multicolumn{5}{c}{\textbf{Overall}}  \\
        & & P & R & F1 & P & R & F1 & Acc & F1-w & F1-M \\
         \midrule
 
            \multirow{13}{*}{\rotatebox[origin=c]{90}{\textbf{IHC}}} & BERTweet &\textbf{91.47 (0.70)} &78.59 
 (2.00) &84.52 (0.87) &60.66 (1.67) &\textbf{81.75 (2.12)} & 69.61 (0.45)  & 79.50 (0.85)  & 80.24 (0.73) & 77.06 (0.63)  \\

 \cmidrule{2-13}        
            &BERTweet+CE  &85.43 (0.75) &89.62 (0.99) &87.47 (0.12) &70.70 (1.16) &62.02 (2.76) & 66.03 (1.16)  & 81.70 (0.16)  & 81.31 (0.31) & 76.75 (0.56)  \\

            &BERTweet+C &90.97 (0.45) &79.77 (0.56) &85.00 (0.17) &61.53 (0.37) &80.34 (1.19) & 69.68 (0.33)  & 79.93 (0.16)  & 80.60 (0.15) & 77.34 (0.18)  \\

            &BERTweet+E &91.22 (0.55) &79.35 (1.29) &84.86 (0.63) &61.27 (1.21) &81.03 (1.51) & 69.76 (0.69)  & 79.83 (0.68)  & 80.53 (0.61) & 77.31 (0.62)  \\

            &BERTweet-AF  &90.71 (0.57) &80.92 (0.80) &85.53 (0.31) &62.64 (0.68) &79.40 (1.54) & 70.02 (0.55)  & 80.48 (0.35)  & 81.08 (0.32) & 77.77 (0.37)  \\

            &BERTweet-MoE &90.35 (0.44) &80.71 (0.77) &85.26 (0.29) &62.14 (0.63) &78.58 (1.25) & 69.39 (0.37)  & 80.10 (0.30)  & 80.70 (0.26) & 77.33 (0.27)  \\

            &BERTweet-SLQ &89.86 (0.97) &81.87 (2.09) &85.66 (0.68) &63.21 (1.92) &77.00 (3.13) & 69.35 (0.36)  & 80.47 (0.62)  & 80.98 (0.45) & 77.51 (0.31)  \\

\cmidrule{2-13}

            &E5 & 90.80 (0.82) & 83.81 (2.03) &  87.15 (0.74) & 66.35 (2.09) & 78.88 (2.56) & 72.01 (0.37) &
            82.39 (0.74) & 82.80 (0.60) & 79.58 (0.51)  \\
            
            &Stella & 88.42 (1.34) & 88.31 (1.67) &  88.34 (0.32) & 71.13 (1.88) & 71.21 (4.26) & 71.07 (1.43) &
            83.39 (0.40) & 83.38 (0.47) & 79.70 (0.73)  \\

            &Jasper & 89.40 (0.80) & \textbf{89.66 (1.42)} & \textbf{89.52 (0.45)} & \textbf{74.22 (2.07)} & 73.58 (2.60) & 73.85 (0.85) &
            \textbf{85.04 (0.52)} & \textbf{85.02 (0.47)} & \textbf{81.68 (0.55)}  \\

            &NV-Embed 
            & 91.22 (0.35) & 85.74 (0.66) &  88.39 (0.22) & 69.20 (0.75) & 79.51 (1.03) & \textbf{73.99 (0.24)} &
            83.95 (0.23) & 84.26 (0.19) & 81.19 (0.19)  \\

\midrule
            \multirow{4}{*}{\rotatebox[origin=c]{90}{\textbf{SBIC}}} 
            & E5 &\textbf{86.43 (0.97)} &82.38 (0.89) &84.35 (0.26) &87.55 (0.46) &\textbf{90.53 (0.88)}&89.01 (0.27)&
            87.09 (0.25) &87.04 (0.24)&86.68 (0.24)  \\
            
            &Stella &85.66 (1.91) &83.77 (2.94) &84.65 (0.73) &88.34 (1.64) &89.67 (1.88)&88.99 (0.27)& 87.18 (0.33)  & 87.16 (0.37) & 86.82 (0.41)  \\

            &Jasper &85.54 (2.01) &\textbf{84.17 (2.99)} &84.80 (0.72) &\textbf{88.61 (1.64)} &89.52 (2.01)&89.03 (0.34)&
            87.26 (0.37)  & 87.24 (0.40) & 86.91 (0.43)  \\

            &NV-Embed &85.78 (0.81) &84.04 (0.62) &\textbf{84.90 (0.15)} &88.51 (0.31) &89.80 (0.74)&\textbf{89.15 (0.23)}&
            \textbf{87.37 (0.20)}  & \textbf{87.35 (0.19)} & \textbf{87.02 (0.18)}  \\

\midrule
            \multirow{4}{*}{\rotatebox[origin=c]{90}{\textbf{DynaHate}}} 
            &E5 &84.61 (0.50) &85.92 (0.80) &85.25 (0.30) &87.64 (0.56) &86.46 (0.61) &87.04 (0.23) &
            86.21 (0.25)  & 86.21 (0.25) & 86.15 (0.25)  \\

            &Stella &87.53 (1.62) &89.44 (2.03) &88.44 (0.25) &90.71 (1.41) &88.91 (1.84)&89.78 (0.27) & 89.16 (0.16)  & 89.16 (0.16) &89.11 (0.16)  \\

            &Jasper &86.50 (1.30) &90.22 (1.84) &88.30 (0.24) &91.23 (1.38) &87.77 (1.62) &89.45 (0.23) &
            88.91 (0.13)  &88.92 (0.13)& 88.88 (0.13) \\

            &NV-Embed & \textbf{88.95 (0.18)}& \textbf{90.64 (0.36)}& \textbf{89.79 (0.17)}&\textbf{91.76 (0.28)}&\textbf{90.25 (0.19)}&\textbf{91.00 (0.13)} &
            \textbf{90.43 (0.15)}  &\textbf{90.44 (0.15)} & \textbf{90.39 (0.15)} \\

\midrule
            \multirow{4}{*}{\rotatebox[origin=c]{90}{\textbf{ToxiGen}}} &E5 &87.32 (0.97) &81.16 (1.51) &84.11 (0.51) &77.34 (1.12) &84.48 (1.62) &80.74 (0.45) &
            82.59 (0.42)  &82.66 (0.41) & 82.43 (0.40)  \\
            
            &Stella &88.71 (0.87)&\textbf{89.95 (1.35)} &\textbf{89.32 (0.58)} &\textbf{86.57 (1.43)} &84.92 (1.44) &\textbf{85.72 (0.68)} &
            \textbf{87.78 (0.61)}  & \textbf{87.76 (0.69)} & \textbf{87.52 (0.61)}  \\

            &Jasper &88.78 (0.92)
            &89.73 (1.35) &89.25 (0.61) &86.33 (1.44) &85.07 (1.52) &85.68 (0.74) &
            87.72 (0.65)  & 87.71 (0.65) & 87.46 (0.66)  \\

            &NV-Embed & \textbf{90.25 (0.90)} & 86.21 (1.26)& 88.18 (0.29) & 82.90 (1.09) &\textbf{87.73 (1.42)} & 85.23 (0.25)&
            86.87 (0.22)  & 86.90 (0.21) & 86.70 (0.21)  \\

\bottomrule
    \end{NiceTabular}
    \label{tab:ihc_in_dataset}
\end{table*}

\noindent\textbf{Implementation details}. All the experiments are conducted on a single NVIDIA H100.
For fine-tuning Stella, Jasper, E5 and BERT-based classifiers, we use a batch size of 16 and AdamW \cite{loshchilov2019decoupledweightdecayregularization} with learning rate $2e^{-6}$ and weight decay $0.5$. We use a linear scheduler with $20\%$ steps of warm up and dropout of 0.2. The models are trained for 4 epochs, and the best one according to the weighted F1 score is selected for the test dataset. For the fine-tuning of NV-Embed, we use LoRA \cite{hu2021loralowrankadaptationlarge} with $r=16$, $\alpha=32$ and dropout of $0.1$. The batch size for fine-tuning NV-Embed is 8. We use a train/test/validation split of 60/20/20 for IHC and DynaHate, 80/10/10 for SBIC and 70/10/20 for ToxiGen.
For linear probing, we use a batch size of 512 and a learning rate of $2e^{-3}$, except for NV-Embed where the batch size is 64 with a learning rate $2e^{-4}$. Training lasts 20 epochs, and the best model according to the weighted F1 score is picked.
For generative models, we use a batch size of 16 with a learning rate of $6e^{-5}$ and the base prompt is the same as the one used for embedding models.

\noindent\textbf{Evaluation protocol}. We use standard classification metrics to evaluate the models' performance: precision, recall, accuracy and F1 scores (weighted and macro). Precision measures the accuracy of positive predictions. High precision is important to avoid over-censorship.
Recall indicates how many of the actual positives are correctly identified by the model. High recall ensures that most of the positive instances from the dataset are detected, which is essential to avoid the spread of hateful speech.
Accuracy measures the overall performance of the model, however, it can be misleading in unbalanced datasets. Therefore, we also consider the F1 score that combines precision and recall.
Weighted F1 is used to overcome class imbalance, avoiding majority class domination, whereas macro F1 gives equal weight to all classes.
We report the mean performance and standard deviation over five runs with different seeds.
We report the models' performance for different metrics to facilitate comparison with existing and future work.
\color{black}
We assess model generalization through cross-dataset evaluation by fine-tuning on IHC or SBIC, respectively, and testing on the held-out datasets.

\noindent \textbf{Note on data contamination}. As recent pre-trained models could have seen IHS datasets in training, we reviewed the training details of the embedding models used in this study and we found no mention of the datasets used in this work. 

\color{black}

\subsection{Enhanced BERT-based classifiers}
Table~\ref{tab:ihc_in_dataset} shows the results of 
BERT-based classifiers under different setups. While
BERTweet alone has the lowest overall accuracy, it outperforms the other variants in  \texttt{not hate} precision and \texttt{hate} recall.
BERTweet+ context gives a slight improvement in most metrics showing that the additional information generated around the tweet helps with the classification.
BERTweet+emotion improves performance across almost all metrics when compared to BERTweet alone. In some cases, such as \texttt{hate} recall, \texttt{hate} F1 score, this version outperforms the model with added context. These improvements show that adding the emotion conveyed by the tweet as a classification feature is useful.
BERTweet+context+emotion gives the highest overall accuracy and weighted F1 score despite showing lower performance in various intra-class metrics.
Adaptive fusion does not give a significant improvement in overall performance metrics over the baseline model, except for a 1 percentage point (p.p.) improvement in accuracy.
This could be due to the fact that each element of the output of the three blocks is already weighted by the MLP input layer.
We observe that the performance of the mixture of experts 
is very similar to that of adaptive fusion, with only minor discrepancies (<1 p.p. variation). This could be due to our implementation of the mixture of experts that scales the outputs of the different blocks.
The shared learnable query case shows similar behavior to adaptive fusion and mixture of experts, with comparable performance and minor variations in the intra-class metrics.

\setlength{\tabcolsep}{4pt}
\renewcommand{\arraystretch}{0.89}
\begin{table*}[h!]
    \centering
     \caption{In-dataset and cross-dataset results for different models trained on IHC~\cite{elsherief-etal-2021-latent} and SBIC~\cite{sap-etal-2020-social} for binary classification with \texttt{hate} as the positive class. We report the average performance across 5 seeds with standard deviation. Models E5, Stella, Jasper and NV-Embed only use the tweet.   * indicates results taken from their corresponding papers. $\dagger$ indicates results taken from related works referencing the method. - indicates results not available in the corresponding papers. For ImpCon~\cite{kim-etal-2022-generalizable}, ShareCon~\cite{ahn-etal-2024-sharedcon} and CCL~\cite{jiang-2025-learn}, we added an extra zero to the results to maintain consistency with other studies that report metrics using two decimal precision. Key- Acc: unweighted accuracy, F1-M: macro F1-score, FT: fine-tuning, LP: linear probing, B: BERT backbone, HB: HateBERT backbone, RB: RoBERTa backbone.}
   
   \begin{NiceTabular}{lcc|cc|cc|cc}
    \toprule
        \textbf{Model} & \multicolumn{2}{c}{\textbf{IHC}}  & \multicolumn{2}{c}{\textbf{SBIC}} & \multicolumn{2}{c}{\textbf{DynaHate}} &  \multicolumn{2}{c}{\textbf{ToxiGen}}\\
         & \multicolumn{2}{c}{in-dataset}  & \multicolumn{2}{c}{cross-dataset} & \multicolumn{2}{c}{cross-dataset} &  \multicolumn{2}{c}{cross-dataset}\\
        &  Acc& F1-M &  Acc & F1-M &    Acc & F1-M &   Acc & F1-M   \\
        
         \midrule
           
            ImpCon* (B)~\cite{kim-etal-2022-generalizable}&  - &78.00 & -&60.70& - & 57.90 &  -& - \\

            ImpCon$^{\dagger}$ (B)~\cite{kim2024labelawarehardnegativesampling} &-& 78.39 &- &54.55 & -& 59.41 & - &59.64 \\
            
             LAHN* (B)~\cite{kim2024labelawarehardnegativesampling} &- &78.62 &- &62.02&- & 56.13 & - &62.92 \\
             
              SharedCon* (B)~\cite{ahn-etal-2024-sharedcon} &- &78.50 &-&65.20&- & 59.50 & - &- \\           

              CCL* (B)~\cite{jiang-2025-learn} &- &78.40 &-&65.30&- & 62.70 & - &- \\    
\midrule
            ImpCon* (HB)~\cite{kim-etal-2022-generalizable} &- &77.40 &-&63.50&- & 59.40& -&- \\
            CCL* (HB)~\cite{jiang-2025-learn} &- &77.60 &-&66.40&- & 63.10 & - &- \\   
\midrule
            ImpCon$^{\dagger}$ (RB)~\cite{kim2024labelawarehardnegativesampling} &-& 78.78 &- &63.82 & -& 50.13 & - &61.79 \\
            
             LAHN* (RB)~\cite{kim2024labelawarehardnegativesampling} &- &80.58 &- &64.01&- & 49.54 & - &64.49 \\
\midrule
              ConPrompt*~\cite{kim-etal-2023-conprompt} &- &77.82 (0.18) &-&67.88 (3.22)&- & 59.28 (0.84)  & - &- \\  

\midrule
              Llama3-8B ~\cite{grattafiori2024llama3herdmodels} &82.35 (0.43) &78.39 (0.50) &61.44 (2.83)& 60.08 (3.09)&57.24 (1.13) & 54.17 (1.59)  & 63.63 (6.35) &63.44 (6.14) \\  

              Gemma-7B ~\cite{gemmateam2024gemmaopenmodelsbased} &81.53 (0.55) &77.76 (0.33) & 65.04 (1.60)&64.49 (1.30)& 57.15 (0.70) & 55.18 (0.67) & 65.08 (2.79) & 64.95 (2.65) \\  

              Qwen3-8B ~\cite{yang2025qwen3technicalreport} &80.33 (0.50) &77.02 (0.35) &64.80 (0.65)&63.95 (0.49)&58.55 (0.62) & 56.65 (0.60) & 71.68 (0.92)& 71.35 (0.86) \\

\midrule
            LP E5 &76.96 (1.75) &72.76 (0.92) &63.14 (5.50) &62.00 (6.73) &62.72 (1.62)&62.39 (1.86) &68.89 (1.88) &67.07 (3.25) \\
            
            LP Stella &82.83 (0.26) &79.15 (0.74) &72.85 (1.17) &72.43 (0.91) &65.23 (1.94)&62.27 (3.62) &75.08 (1.44) &74.57 (1.45) \\

            LP Jasper &82.01 (0.53) &78.32 (0.41) &72.40 (1.80) &72.02 (1.48) &64.86 (1.04)&62.14 (2.46) &75.40 (0.65) &74.93 (0.75) \\

            LP NV-Embed &83.78 (0.38) &79.83 (0.40) &73.07 (1.18) &72.68 (0.85) &67.14 (1.81)&65.61 (3.15) &77.29 (1.08) &76.57 (1.14) \\
\midrule           
            FT E5 & 82.39 (0.74) & 79.58 (0.51) & 67.92 (1.84) & 72.60 (1.35) &63.14 (0.57) &58.93 (1.43) & 71.34 (1.78) &71.25 (1.74)  \\
            
            FT Stella & 83.39 (0.40) & 79.70 (0.73) & 73.48 (0.96) & 72.60 (1.35) &66.98 (1.33) &63.21 (2.02) & 80.25 (1.29) &80.16 (1.20)  \\

            FT Jasper & \textbf{85.04 (0.52)} & \textbf{81.68 (0.55)} & 74.55 (1.31) & 73.72 (1.61) &67.59 (1.23) &63.90 (1.86) & 80.85 (1.15) &80.77 (1.11)  \\

            FT NV-Embed & 83.95 (0.23) & 81.19 (0.19) & \textbf{77.24 (0.34)} & \textbf{76.96 (0.31)} &7\textbf{3.59 (0.90)} &\textbf{72.62 (1.27)} & \textbf{85.12 (0.40)} &\textbf{84.84 (0.43)}  \\
            
    \bottomrule
    \end{NiceTabular}    
    \begin{adjustbox}{margin=2.65pt 0pt 0pt 0pt} 
    \begin{NiceTabular}{lcc|cc|cc|cc}
    \toprule
        \textbf{Model} & \multicolumn{2}{c}{\textbf{SBIC}}  & \multicolumn{2}{c}{\textbf{IHC}} & \multicolumn{2}{c}{\textbf{DynaHate}} &  \multicolumn{2}{c}{\textbf{ToxiGen}}\\
         & \multicolumn{2}{c}{in-dataset}  & \multicolumn{2}{c}{cross-dataset} & \multicolumn{2}{c}{cross-dataset} &  \multicolumn{2}{c}{cross-dataset}\\
        &  Acc& F1-M &  Acc & F1-M &    Acc & F1-M &   Acc & F1-M   \\
         \midrule
           
            ImpCon* (B)~\cite{kim-etal-2022-generalizable} &- &83.60 &-&61.40&- & 61.20 & -&- \\
            
            ImpCon$^{\dagger}$ (B)~\cite{kim2024labelawarehardnegativesampling} &-& 83.53 &- &58.64 & -& 59.50 & - &66.54 \\

            LAHN* (B)~\cite{kim2024labelawarehardnegativesampling} &- &84.31 &- &61.58&- &60.97 & - &68.52\\

            SharedCon* (B)~\cite{ahn-etal-2024-sharedcon} &- &84.30 &-&62.40&- & 62.00 & - &- \\           

            CCL* (B)~\cite{jiang-2025-learn} &- &84.30&-&61.30&-& 62.10 & - &- \\  

\midrule
            ImpCon* (HB)~\cite{kim-etal-2022-generalizable} &- &84.80 &-&59.90&- & 60.60& -&- \\
            CCL* (HB)~\cite{jiang-2025-learn} &- &84.80 &-&61.50&- & 61.90 &- &- \\   
\midrule   

            ImpCon$^{\dagger}$ (RB)~\cite{kim2024labelawarehardnegativesampling} &-& 84.66 &- &56.95 & -& 60.70 & - &66.77 \\
            
             LAHN* (RB)~\cite{kim2024labelawarehardnegativesampling} &- &85.80&- &64.05&- & 63.26 & - &69.91\\

\midrule
              ConPrompt*~\cite{kim-etal-2023-conprompt} &- &\textbf{88.85 (0.23)} &-&66.27 (0.44)&-& 67.59 (0.64)  & - &- \\  

              Fr-HARE*~\cite{yang2023hare} &85.21 &-&-&-&68.06 & -  & - &- \\  

              CO-HARE*~\cite{yang2023hare} &84.93 &-&-&-&69.98 & -  & - &- \\  

\midrule
            LP E5 &81.65 (0.40) &80.92 (0.43) &52.59 (1.83) &52.55 (1.78) &63.37 (1.24)&59.23 (2.55) &65.38 (1.69) &64.98 (1.99) \\
            
            LP Stella &86.05 (0.09) &85.69 (0.10) &63.40 (0.57) &62.54 (0.44) & 68.79 (0.48)&66.49 (0.78) &72.95 (1.05) &72.94 (1.05) \\

            LP Jasper &85.81 (0.15) &85.44 (0.19) &64.29 (1.40) &63.25 (1.04) &67.96 (0.60)&65.54 (1.14) &72.95 (0.86) &72.93 (0.89) \\

            LP NV-Embed &85.96 (0.29) &85.62 (0.24) &64.52 (1.18) &63.59 (0.89) &68.79 (0.96)&66.64 (1.62) &77.02 (1.45) &76.98 (1.40) \\
 \midrule
            FT E5 &87.09 (0.25) &86.68 (0.24) &59.19 (0.90) &58.72 (0.76) &66.85 (0.61)&63.84 (1.03) &75.12 (1.35) &75.09 (1.29) \\
            
            FT Stella &87.18 (0.33)  &86.82 (0.41) &66.75 (2.75) &65.44 (2.22) &70.97 (1.94)&68.66 (3.02)&80.57 (3.53) &80.46 (3.50) \\

            FT Jasper &87.26 (0.37)  &86.91 (0.43) &66.92 (2.65) &65.52 (2.07) &71.10 (1.96)
            &68.89 (2.98)
            &81.06 (3.90) &80.94 (3.84) \\

            FT NV-Embed &\textbf{87.37 (0.20)} &87.02 (0.18) &\textbf{67.52 (0.70)} &\textbf{66.38 (0.54)} &\textbf{72.55 (0.46)}&\textbf{70.51 (0.62)} &\textbf{84.08 (0.34)} &\textbf{84.00 (0.32)} \\
        
    \bottomrule
    \end{NiceTabular}
    \end{adjustbox}
    \label{tab:ihc_cross}   

\end{table*}

\color{black}
\subsection{Specialized generalist embeddings}

\noindent\textbf{In-dataset evaluation}.
For {\em in-dataset} evaluation (see Tables \ref{tab:ihc_in_dataset} and \ref{tab:ihc_cross}), we get 1.1 p.p. improvement over LAHN \cite{kim2024labelawarehardnegativesampling} on IHC, and lack 1.83 p.p. compared to ConPrompt \cite{kim-etal-2023-conprompt} on SBIC. Interestingly, fine-tuning a larger model like NV-Embed is not always the optimal choice when evaluating F1-macro scores. NV-Embed achieves the best performance only on two datasets: SBIC and DynaHate. In terms of performance on IHC and ToxiGen, Jasper and Stella are the best models when fine-tuned. Linear probing is less effective than fine-tuning, but the trade-off between fine-tuning a smaller model, such as E5, or using linear probing on NV-Embed is not trivial. Results and analysis for in-dataset linear probing with E5, Stella, Jasper and NV-Embed are reported in Appendix \ref{app:in_dataset_linear_probing}.

\begin{figure}[t!]
    \centering

    \includegraphics[width=1\columnwidth]{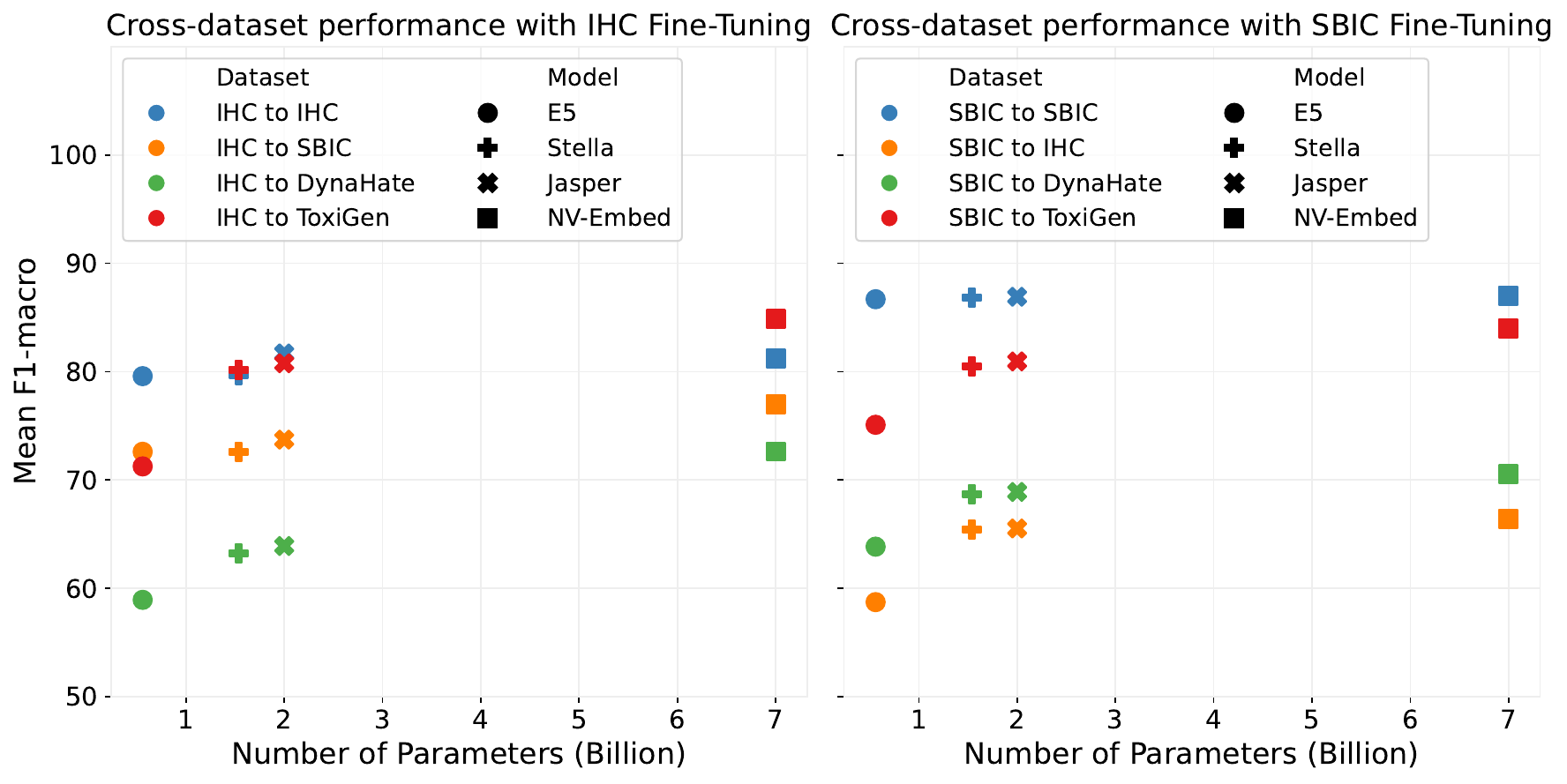}

    \caption{F1-macro scores for cross-dataset evaluation, averaged over 5 seeds,  and different model sizes: larger models achieve higher performance. }
    \Description{Two side-by-side scatter plots showing cross-dataset performance in mean F1-macro score versus model size in billions of parameters. The left plot uses IHC fine-tuning, and the right uses SBIC fine-tuning. Each point is colored by the source–target dataset pair and shaped by model type (E5, Stella, Jasper, NV-Embed). In both plots, performance varies with model size and dataset pairing.}
    \label{fig:cross-dataset-parameters}
\end{figure}

\noindent\textbf{Cross-dataset evaluation}.
On {\em cross-dataset} testing, we observe that the bigger the model, the better it performs. In Table \ref{tab:ihc_cross}, we see that general text embedding models fine-tuned on IHC outperform all previous work, except for E5 which loses -0.35 p.p. in F1-macro compared to ConPrompt. IHC appears to be the best training dataset for generalization, when using NV-Embed with a substantial 20.35 p.p. improvement in macro F1 over LAHN~\cite{kim2024labelawarehardnegativesampling} on ToxiGen. Stella, Jasper and NV-Embed also prove to be well-performing with linear probing in the cross-dataset setting when being trained on IHC, as they all surpass previous results on this task. The results for cross-dataset evaluation after fine-tuning on SBIC (Table~\ref{tab:ihc_cross}) are noteworthy, even if they are not as impressive in certain cases. Nonetheless, we achieve state-of-the-art performance in cross-dataset evaluation using NV-Embed. Linear probing outperforms the previous state-of-the-art model, LAHN~\cite{kim2024labelawarehardnegativesampling}, only on the ToxiGen dataset.
Using NV-Embed with linear probing proves to be more effective than a fine-tuned E5. It is interesting to note that linear probing in the cross-dataset setting is very detrimental for NV-Embed. In an in-dataset setting, the performance drop is less than 2 p.p., while in a cross-dataset setting, the drop can go up to 7 p.p.. The choice between using linear probing on a larger model or fine-tuning a smaller model depends on the specific requirements. A fine-tuned smaller model offers higher throughput, while a linear probing approach on NV-Embed provides better performance in the cross-dataset setting. Figure~\ref{fig:cross-dataset-parameters} shows the cross-dataset performance variations with relation to the number of parameters in each model (see Appendix \ref{app:appendix_Additional_results} for in-dataset performance 
comparison). An overview of the computational requirements is provided in Appendix \ref{app:appendix_computational_requirements}.

\noindent\textbf{Comparison with generative models}.
Our experiments with Llama3-8B, Gemma-7B and  Qwen3-8B show no improvements over state-of-the-art methods. 
LLMs are not designed for classification and alignment impacts their performance on hate speech content~\cite{lee2025nvembedimprovedtechniquestraining}.
Using Qwen3-8B on IHC intra-dataset gives 80.33 accuracy and 77.02 F1-macro. Cross-dataset evaluation IHC-SBIC gives 64.80 accuracy and 63.95 F1-macro, while IHC-ToxiGen gives 71.68  accuracy and 71.35  F1-macro.
These results underscore the limitations of current instruction-tuned LLMs for hate speech detection, highlighting the need for models explicitly optimized for this classification task.

\section{Analysis}
\noindent \textbf{Confidently misclassified samples}. 
We examine IHC~\cite{elsherief-etal-2021-latent} samples misclassified with high confidence in its class assignment by extracting the model's predictions along with their probabilities.
We first look at the examples for which BERTweet+context+emotion is confident that a \texttt{not hate} sample is a \texttt{hate} one. An example is: \textit{"in other words "we need to keep brainwahsing whites to accept their destruction." \#istandfordiversity"}~\cite{elsherief-etal-2021-latent}. These examples are labeled as \texttt{not hate} in the dataset, but the ambiguous and implicit nature of the samples makes them hard to classify even for humans, which can lead to annotation errors. Most of the confidently misclassified samples are ambiguous.
We notice that the model is very confident and correct when it comes to \texttt{hate} examples that contain hate toward immigrants, especially when tweets call for them to be sent back to their countries of origin, for example: \textit{"they need to stay hidden in the shadows, otherwise, deport them as the law requires !"}~\cite{elsherief-etal-2021-latent}.
By looking at the misclassified \texttt{hate} samples, we see real examples of \texttt{implicit hate} where the model fails to understand it. These heavily rely on a context or external knowledge, for example \textit{"most people were ok with rebel battle flag confederate"}~\cite{elsherief-etal-2021-latent}.
Similarly to the BERTweet-based model we develop, NV-Embed sometimes misclassifies \texttt{not hate} samples in the \texttt{hate} class,
 for example \textit{"it didn’t take long to figure out [...] were/are behind white-guilt concept that’s ravaging western-societies"}~\cite{elsherief-etal-2021-latent}. The dataset annotation might also be the cause for this, since some are obviously hateful.
We notice again a tendency to correctly classify \texttt{hate} samples about sending immigrants back as \texttt{hate} with high confidence, for example \textit{"round them up \& deport them."}~\cite{elsherief-etal-2021-latent}.
By looking at the examples of \texttt{hate} samples being classified incorrectly, we see that these samples seem to be misclassified because of a lack of background knowledge of the model. For example, the sample \textit{"he is ranting because the alt-lite has collapsed . the alt-right is being proven right about nationalism"}~\cite{elsherief-etal-2021-latent} requires knowledge of what the alt-right is.
The model seems to be more confident and correctly classifies \texttt{not hate} samples when these tweets state information or give news about something that happened like \textit{"more : the russian bombers will reportedly launch from the ’engels’ airbase and will be armed with cruise missiles "}~\cite{elsherief-etal-2021-latent}.
See Tables ~\ref{tab:misclassified_samples_base_model} and ~\ref{tab:misclassified_samples_nvembed} of Appendix \ref{app:appendix_missclass_samples} for more examples.

\noindent \textbf{Topic analysis}. We perform topic modeling on hate speech data from IHC~\cite{elsherief-etal-2021-latent} to identify the most prevalent themes and key patterns that characterize hate speech. 
As a pre-processing step, we remove stop words using the NLTK~\footnote{Natural Language Toolkit, https://www.nltk.org/} dictionary and then pass the cleaned textual data to the BERTopic~\cite{grootendorst2022bertopic} model.
We analyze the test dataset to extract its topics and report the most frequent ones in Table \ref{tab:most_common_topics}. 
%


%
We observe a significant bias toward racism and black hate in the data. To further understand the misclassified topics, we analyze the topics in the misclassified samples using the same procedure applied to the entire test set.  
We notice that some topics present in the test set do not appear as much in the misclassified samples, for example, the topics about India or gay marriage. 

By comparing the predictions NV-Embed-based classifier for both classes, we can see that certain topics are more often misclassified when part of a certain class. 
For example, the model tends to classify unharmful tweets about immigration or Jewish people as \texttt{hate}.
Discussion about US right/altright tend to be more classified as \texttt{not hate}, even when hateful.
Some topics seem to be balanced between classification mistakes in both classes: for example, the tweets about Islam are misclassified in both ways. See Appendix \ref{app:appendix_additional_analysis} 
for visualizations of the main misclassified topics and sample distribution within the topics.

\begin{table}[t!]
\centering
\small
\caption{Words representing the most common topics in the test set of Implicit Hate Corpus dataset (some synonyms/repeated variations of words were removed manually from the representations to enhance readability). Shown words are the most frequently occurring words within topic clusters, the group of most representative words forms the representation of the topic. The Count column shows the number of tweets per topic. Words may overlap across topics, for example, the first and second topics show racial terms, but differ in focus: general hostility versus a political context. }
\label{tab:most_common_topics}
\begin{NiceTabular}{lcc}
\toprule
{\bf Count} & {\bf Representation} \\ \midrule
1315 & white, people, racist, race, black, hate \\
915  &white, racist, black, america, supremacists, nationalism \\
397  &  jews, islam, muslims, religious, islamic, israel, kill \\
338  & antifa, altright, house, media, right, trump, populist  \\
212  & illegals, wall, deport, border, laws, immigrants \\
155  & india, delhi, hindus, bjp, indian, modi \\
101  & marriage, abortion, parenthood, prolife, gay, unborn, kill\\
67    &  holocaust, hitler, news, adolf, germans, denial \\
37    &  cruz, ted, heidi, trump, rubio, texas, nomination, vote  \\ \bottomrule
\end{NiceTabular}
\end{table}
\noindent\textbf{Target bias}. We observe that the NV-Embed based classifier shows sensitivity to the text target, with higher \texttt{hate} probabilities for statements like 
\textit{Black people are stupid} (0.65), \textit{White people are stupid} (0.53), \textit{Jews are stupid (0.69)}, \textit{Muslims are stupid} (0.68), and \textit{Gay are stupid} (0.55) while vague phrases like \textit{They are stupid} (0.41) have lower scores. Similarly, text like \textit{White people are superior} (0.69), \textit{Muslims are superior} (0.71), \textit{The state should refuse entry to white people} (0.61) and \textit{The state should refuse entry to Jews} (0.70) also show higher \texttt{hate} probabilities, highlighting the model's sensitivity to specific targets and wording.

\section{Conclusion}
\label{sec:conclusion}

We present new state-of-the-art benchmarks for IHS detection by leveraging recent pre-trained LLM-based embeddings.  We show that adding specialized components for hate speech detection using BERT as backbone provides only minimal improvements. Instead, refining recent LLM-based embedding models provides substantial improvements in generalization across datasets.  We discuss a comprehensive evaluation on several datasets and identify the models that better capture the underlying patterns and nuances of IHS. 
As future work, the detection of subtle and implicit forms of hate speech could be enhanced by exploring visual augmentation with a diffusion model to generate images from text~\cite{yang2024trustworthyhatespeechdetection}.
Another direction could involve assessing the detection capabilities in multilingual settings.

\bibliographystyle{ACM-Reference-Format}
\balance
\bibliography{main}

\newpage
\appendix
\nobalance
\pagestyle{plain}
\pagenumbering{gobble} 
\twocolumn[
\begin{center}
    {\LARGE \bfseries Appendices for "Specializing General-purpose LLM Embeddings for Implicit Hate Speech Detection across Datasets"
    \par}

\end{center}
\vspace{0.75cm}
]

\section{Context generation with Llama2}
\label{app:appendix_context generation}

Some examples of tweets and corresponding generated context with uncensored Llama2\footnote{\url{https://huggingface.co/georgesung/llama2_7b_chat_uncensored}}  are provided in Table \ref{tab:generated_context}. As we can see, the LLM explains the context, but also gives an interpretation of the tweet.

\begin{table*}[h]
\caption{Examples of tweets from IHC~\cite{elsherief-etal-2021-latent} and their corresponding context generated by Llama 2.}
\begin{NiceTabular}{p{5cm} | p{12cm}}
\toprule
\textbf{Tweet} & \textbf{Generated context} \\ \midrule

This may be a horrible thing to say but are black lives matter \#blm the new ku klux klan? They both racially hate \& despise others &
The tweet suggests that Black Lives Matter (BLM) is similar to the Ku Klux Klan (KKK) in terms of racial hatred and despising others. It may be implying that both groups promote violence and discrimination against certain groups. However, it's important to note that BLM is a civil rights movement addressing systemic racism and police brutality towards Black people, while the KKK is a white supremacist organization that has historically oppressed Black people. The tweet may also be intended as sarcasm or satire. \\ \midrule

stupid white liberals at a demonstration  apologizing for slavery of blacks ... words fail me ...&
The tweet appears to be a criticism of white liberals who are apologizing for the slavery of black people. The use of the word "stupid" suggests that the author believes that the liberals are making a foolish or ignorant statement. The use of the word "apologizing" suggests that the liberals are taking responsibility for something they did or did not do. The use of the word "demonstration" suggests that the liberals are participating in a public event or protest. The use of the word "fail" suggests that the author is unable to comprehend or accept the liberals' apology. The use of the word "blacks" suggests that the author is referring to a specific group of people. The use of the word "words" suggests that the author is unable to find the appropriate words to express their feelings or opinions. Overall, the tweet appears to be a criticism of white liberals who are apologizing.\\ \bottomrule
\end{NiceTabular}
\label{tab:generated_context}
\end{table*}

\section{In-Dataset linear probing}
\label{app:in_dataset_linear_probing}

Table~\ref{tab:IHC_in_dataset_LP} shows the results of E5, Stella, Jasper and NV-Embed with linear probing on all 4 used datasets.
On ToxiGen, linear probing on NV-Embed is very effective and achieves approximately $\mathbf{3}$ more percentage points in the F1-macro score compared to a fine-tuned E5. On the IHC dataset, the performance of linear probing on NV-Embed and fine-tuning on E5 is similar. However, on DynaHate and SBIC, a fine-tuned E5 outperforms NV-Embed with linear probing. It is also worth mentioning that on ToxiGen, NV-Embed with linear probing is only losing $\mathbf{2}$ percentage points in F1-macro compared to its fine-tuned version. 

\setlength{\tabcolsep}{5.5pt}
\begin{table*}[h!]
    \centering
    \small
    \caption{Results with linear probing on IHC~\cite{elsherief-etal-2021-latent}, SBIC~\cite{sap-etal-2020-social}, Dynahate~\cite{vidgen2021learning} and ToxiGen~\cite{hartvigsen-etal-2022-toxigen} datasets for binary classification with \texttt{hate} as the positive class. We report the average over 5 runs with different seeds, and the standard deviation for each metric is in parentheses. Models E5, Stella, Jasper and NV-Embed only use the tweet. Key- Acc: unweighted accuracy, P: precision, R: recall, F1-w: weighted F1-score, F1-M: macro F1-score, LP: linear probing.}
    \begin{NiceTabular}{clccc|ccc|ccccc}
        \CodeBefore
            \cellcolor[gray]{0.9}{3-1,4-1,5-1,6-1, 7-1, 8-1, 9-1, 10-1, 11-1,12-1,13-1, 14-1, 15-1, 16-1, 17-1, 18-1, 19-1}
        \Body
    \toprule
        & \textbf{Model} & \multicolumn{3}{c}{\texttt{\textbf{Not hate}}} & \multicolumn{3}{c}{\texttt{\textbf{Hate}}} & \multicolumn{5}{c}{\textbf{Overall}}  \\
        & & P & R & F1 & P & R & F1 & Acc & F1-w & F1-M \\
         \midrule
 
            \multirow{4}{*}{\rotatebox[origin=c]{90}{\textbf{IHC}}} 
            &LP E5 &86.01 (2.75)&81.13 (6.23) &83.30 (2.14) &59.47 (4.38) &66.60 (9.90) &62.23 (2.25)& 76.96 (1.75)  & 77.25 (1.20) & 72.76 (0.92)  \\

            &LP Stella &88.40 (1.60) &87.45 (2.10) &87.89 (0.32) &69.78 (2.14) &71.38 (5.23) &70.42 (1.64)& 82.83 (0.26)  & 82.88 (0.38) & 79.15 (0.74)  \\

            &LP Jasper &88.09 (1.28) &86.46 (1.28) &87.26 (0.49) & 67.92 (1.59) &70.97 (1.62) &69.38 (0.46) & 82.01 (0.53)  & 82.13 (0.43) & 78.32 (0.41)  \\

            &LP NV-Embed &87.83 (1.30) &89.73 (2.22) &88.74 (0.47) &73.27 (2.90) &69.01 (4.47) &70.92 (1.04) & 83.78 (0.38)  & 83.63 (0.25) & 79.83 (0.40)  \\
\midrule
            \multirow{4}{*}{\rotatebox[origin=c]{90}{\textbf{SBIC}}} 
            &LP E5 &81.19 (2.97) &73.92 (4.18) &77.25 (0.99) &82.16 (1.91) &87.30 (3.21) &84.59 (0.64)& 81.65 (0.40)  & 81.49 (0.39) & 80.92 (0.43)  \\

            &LP Stella &83.82 (0.95) &83.02 (1.40) &83.41 (0.25) &87.69 (0.78) &88.27 (1.02) &87.97 (0.15) & 86.05 (0.09)  & 86.04 (0.09) & 85.69 (0.10)  \\

            &LP Jasper &83.46 (0.96) &82.85 (1.59) &83.14 (0.37) &87.54 (0.89) &87.97 (1.06) &87.75 (0.14) & 85.81 (0.15)  & 85.80 (0.17) & 85.44 (0.19)  \\

            &LP NV-Embed &83.95 (1.33) &83.42 (1.18) &83.39 (0.13) &87.88 (0.60) &87.83 (1.34) &87.84 (0.38) & 85.96 (0.29)  & 85.96 (0.26) & 85.62 (0.24)  \\

\midrule
            \multirow{4}{*}{\rotatebox[origin=c]{90}{\textbf{DynaHate}}} 
            
            &LP E5 &76.60 (1.88) &68.31 (2.34) &72.17 (0.53) &74.92 (0.83) &81.84 (2.54) &78.20 (0.75) & 75.56 (0.36)  & 75.40 (0.29) & 75.18 (0.27)  \\

            &LP Stella &80.81 (0.77) &83.71 (1.05) &82.22 (0.21) &85.45 (0.65) &82.77 (1.06) &84.02 (0.29) & 83.20 (0.19)  & 83.22 (0.19) & 83.15 (0.19)  \\

            &LP Jasper &80.18 (0.29) &82.64 (0.45) &81.39 (0.11) &84.56 (0.28) &82.31 (0.41)&83.42 (0.10) & 82.46 (0.08) & 82.48 (0.08) & 82.41 (0.08) \\

            &LP NV-Embed &81.25 (0.81) &83.95 (1.69)&82.56 (0.44) &85.71 (1.12) &83.20 (1.22) &84.42 (0.18) & 83.55 (0.21)  & 83.56( 0.22) & 83.49 (0.23)  \\
\midrule
            \multirow{4}{*}{\rotatebox[origin=c]{90}{\textbf{ToxiGen}}} 
            &LP E5 &85.41 (0.47) &79.10 (2.32) &82.11 (1.11) &74.99 (1.80) &82.21 (1.12) &78.41 (0.60) & 80.44 (0.90)  & 80.52 (0.88) & 80.26 (0.85)  \\

            &LP Stella &86.80 (0.43) &85.24 (0.92) &86.01 (0.29) & 81.05 (0.83)& 82.95 (0.82) & 81.98 (0.17) & 84.25 (0.23)  & 84.27 (0.21) & 84.00 (0.20)  \\

            &LP Jasper &86.66 (0.25) &84.19 (0.59) &85.40 (0.21)&79.96 (0.51) & 82.95 (0.47) &81.43 (0.14) & 83.65 (0.18)  & 83.69 (0.17) & 83.42 (0.17)  \\

            &LP NV-Embed &85.31 (1.26) &90.93 (2.22)&88.01 (0.77) &87.04 (2.54) &79.35 (0.25) &82.97 (0.92) & 85.93 (0.77)  & 85.83 (0.76) & 85.49 (0.77) \\

\bottomrule
    \end{NiceTabular}
    \label{tab:IHC_in_dataset_LP}
\end{table*}

\section{Additional results}
\label{app:appendix_Additional_results}
\begin{figure*}[h]
    \centering
    \includegraphics[width=0.7\columnwidth]{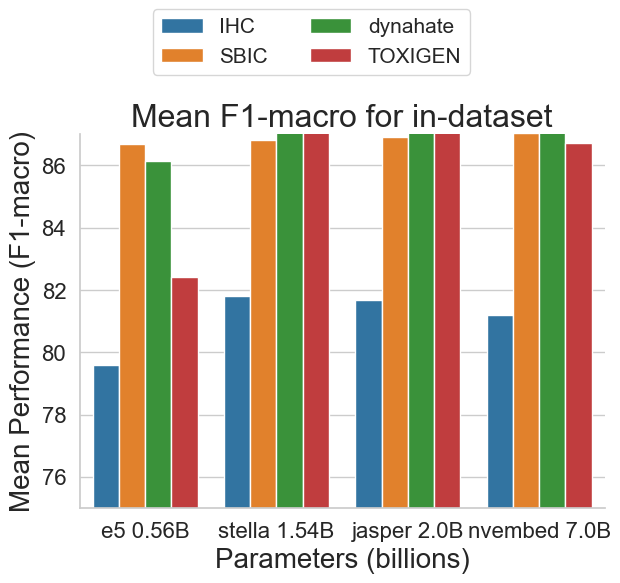}
    \caption{Performances in F1-macro of in-dataset evaluation averaged over 5 seeds with different model sizes. } 
    \label{fig:in-dataset-parameters}
    \Description{A bar plot showing in-dataset performance in mean F1-macro score versus model size in billions of parameters. }
\end{figure*}

Figure~\ref{fig:in-dataset-parameters} shows in-dataset performance in F1-macro for comparison between different model sizes.

\section{Computational requirements}
\label{app:appendix_computational_requirements}

Table~\ref{tab:model_performance} reports the mean samples per second and GPU requirements.  These numbers were obtained with a batch of 16 for E5, Stella and Jasper and a batch of 8 for NV-Embed. The batch size can be scaled up to fill up the GPU at deployment time. Previous SOTA methods rely on BERT based methods which are 3 times smaller than E5. However, we see that E5 gives high throughput. E5 can be loaded on a consumer grade GPU.

\begin{table*}[t!]
\centering
\caption{Mean processing inference speed and GPU memory usage. These numbers are obtained using the specified testing configuration.}
\begin{NiceTabular}{l c c}
\toprule
\textbf{Model} & \textbf{Samples/sec (mean)} & \textbf{GPU Memory (GB)} \\
\midrule
E5       & 1225 & 12.0 \\
Stella   & 714  & 35.2 \\
Jasper   & 616  & 36.0 \\
NV-Embed  & 184  & 62.4 \\
\bottomrule
\end{NiceTabular}
\label{tab:model_performance}
\end{table*}

\begin{table*}[!h]
\small
\caption{Samples from IHC~\cite{elsherief-etal-2021-latent}  classified with high confidence by BERTweet with context and emotion. }
\begin{NiceTabular}{lcc}
\toprule
    \textbf{Prediction} & \multicolumn{2}{c}{\textbf{Correct label}} \\ 
     \multirow{10}{*}{\rotatebox[origin=c]{90}{\texttt{hate}}}&  \texttt{hate} & \texttt{not hate}  \\ \hline 
         & \multicolumn{1}{l|}{
    \begin{tabular}[c]{@{}l@{}}
    - as long as white men allow jews into their midst\\
    and to lead them they will continue to fail their race .\\
    - they need to stay hidden in the shadows, otherwise,\\
   deport them as the law requires !\\
    - stop ' warehousing ' them and send them back\\
    where they came from ! no visa , no stay--gone ! !
    \end{tabular}} &    
    
    \begin{tabular}[c]{@{}l@{}}
    - it 's not fair for black americans to go jobless well you\\
    import illegals to replace them !\\
    - it didn't take long to figure out jews were / are behind\\
    white-guilt concept that 's ravaging western-societies- \\
    really ? so the jews are masterminding immigration\\
    policies in order to wipe out white people ?\\ 
    - in other words "we need to keep brainwahsing whites\\ 
    to accept their destruction." \#istandfordiversity
    \end{tabular} \\ \hline 
    
   \multirow{-2}{*}{\rotatebox[origin=c]{90}{\texttt{not hate}}} & \multicolumn{1}{l|}{
    \begin{tabular}[c]{@{}l@{}}
    - its cowardice that bannon denounced the alt right.\\ 
    i no longer support him\\ 
    - most people were ok with rebel battle flag confederate\\ 
    - italian authorities have blocked access to stormfront\\ 
    in italy and arrested patriots for posting on it !\end{tabular}}  & 
    
    \begin{tabular}[c]{@{}l@{}}
    - more relatives of \#chinas top leaders implicated in\\ 
    \#panamapapers\\ 
    - poll : trump dominates in nevada south carolina overall\\ 
    and on the issues\\ 
    - collective exhale after high court announcement \\ 
    \#neilgorsuch
    \end{tabular}      \\
     \hline                                                       
    \end{NiceTabular} 
\label{tab:misclassified_samples_base_model}
\end{table*}

\begin{table*}[t]
\small
\caption{Samples from IHC~\cite{elsherief-etal-2021-latent}  classified with high confidence by NV-Embed. }
\begin{tabular}{lcc}
\toprule
    \textbf{Prediction} & \multicolumn{2}{c}{\textbf{Correct label}} \\ 
     \multirow{10}{*}{\rotatebox[origin=c]{90}{{\texttt{hate}}}}&  \texttt{hate} & \texttt{not hate}  \\ \cline{1-3} 
         & \multicolumn{1}{l|}{
    \begin{tabular}[c]{@{}l@{}}
    - no one cares, more illegals making illegal entry\\
    into our nation. like any criminal, some got theirs.\\
    deport the rest.\\ 
    - round them up \& deport them.\\ 
    - yep, the deal is, they get deported and so do their\\
    illegal parents. then we build the wall so they never\\
    come back. that's my dream. does that make me a\\
    dreamer?
    \end{tabular}}&
    \begin{tabular}[c]{@{}l@{}}
    - it's not fair for black americans to go jobless well\\
    you import illegals to replace them !\\ 
    - it didn't take long to figure out jews were / are behind\\
    white-guilt concept that's ravaging western-societies\\ 
    - blacks \& latinos attack \& kill white people daily but\\
    when blacks or latinos attack \& kill cops it's a big deal?\\
    just sayin
    \end{tabular}   \\ 
    \hline
    
   \multirow{-2}{*}{\rotatebox[origin=c]{90}{\texttt{not hate}}} & \multicolumn{1}{l|}{
    \begin{tabular}[c]{@{}l@{}}
    - \#macon ga : macon's mlk drive ebt marts are wrapped\\
    in anti-theft caging. and tacky yellow anti-theft cages\\
    at that.\\ 
    - cuckservatives : yes the alt-right are just a bunch of\\
    racists\\ 
    - he is ranting because the alt-lite has collapsed .\\
    the alt-right is being proven right about \\
    nationalism
    \end{tabular}} &
    
    \begin{tabular}[c]{@{}l@{}}
        - shri ajay tamta wins in almora\\ 
    - more : the russian bombers will reportedly launch\\
    from the 'engels' airbase and will be armed with cruise\\
    missiles .\\ 
    - this piece seems to conflate 2 positions . i believe\\
    royce will lead hhs faith-based office but not overall\\
    administration faith-based office
    \end{tabular} \\
    \hline 
    \vspace{1cm}
 \end{tabular}
\label{tab:misclassified_samples_nvembed}
\end{table*}

\section{Misclassified samples with high confidence}
\label{app:appendix_missclass_samples}
Tables~\ref{tab:misclassified_samples_base_model} and~\ref{tab:misclassified_samples_nvembed} show examples of correctly and wrongly classified samples, for which BERTweet and NV-Embed output a high confidence.

\section{Additional analysis}
\label{app:appendix_additional_analysis}
\begin{figure*}[h]
    \centering    
    \includegraphics[width=0.8\columnwidth]{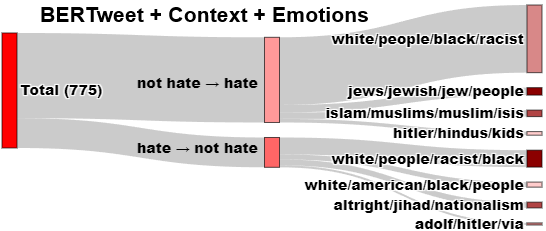}    \includegraphics[width=0.8\columnwidth]{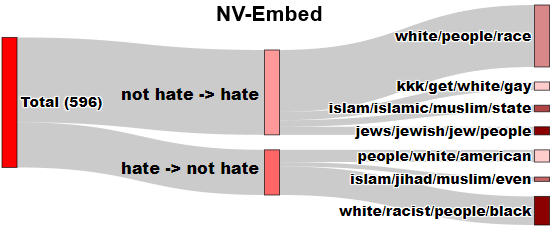}
    \caption{Misclassification distribution of IHC~\cite{elsherief-etal-2021-latent} test samples and their recurrent topics. Each Sankey diagram represents misclassified samples from the test set with a model (left: BERTweet+context+emotion, right: NV-Embed). The middle nodes represent the type of misclassification (\texttt{not hate} classified as \texttt{hate} and vice versa). The right nodes show the topics of the misclassified samples extracted with BERTopic.}
    \Description{Two Sankey diagrams comparing misclassification flows for IHC test samples using different BERTweet+context+emotion and NV-Embed models, showing the number of samples, mispredicted classes and the labels of associated topic phrases.}
    \label{fig:topic_diagram}
\end{figure*}

\begin{figure*}[t]
    \centering

    \includegraphics[width=0.9\columnwidth]{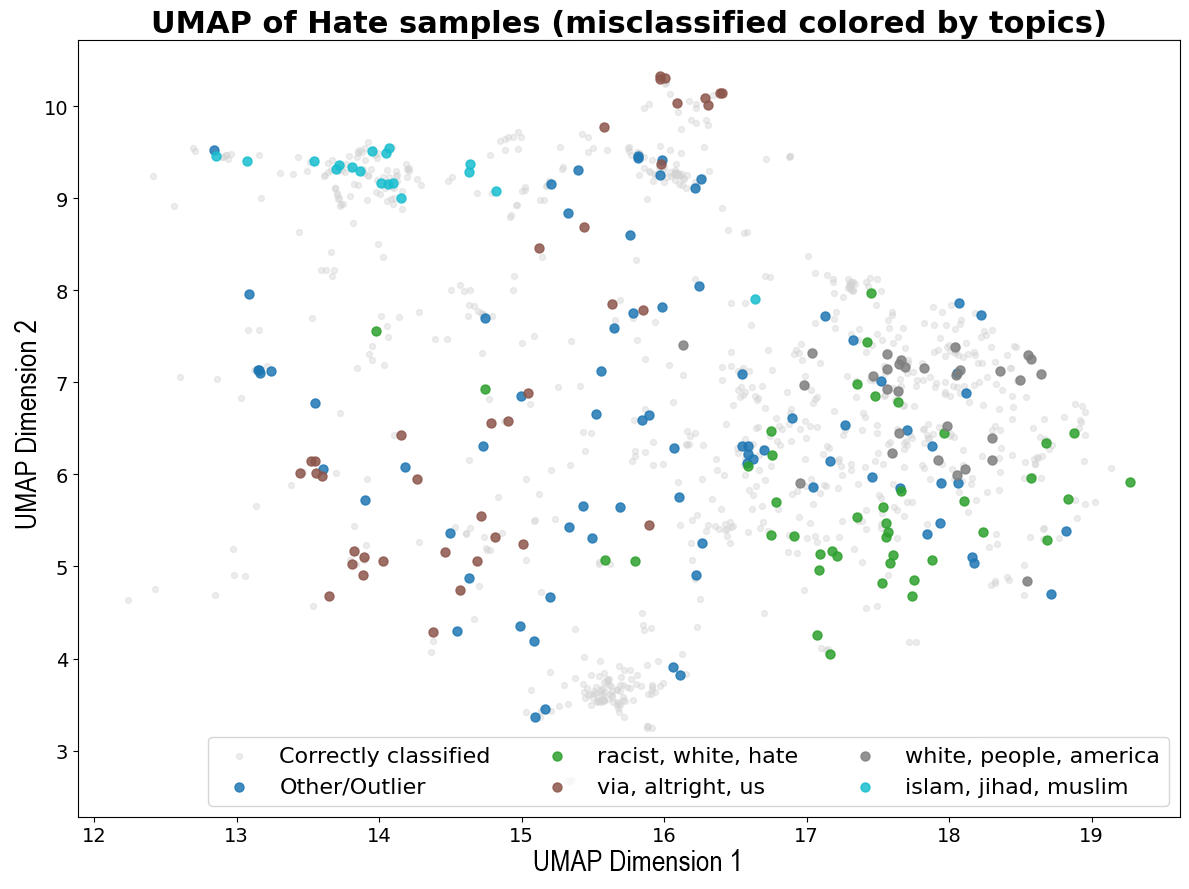}
    \includegraphics[width=0.89\columnwidth]{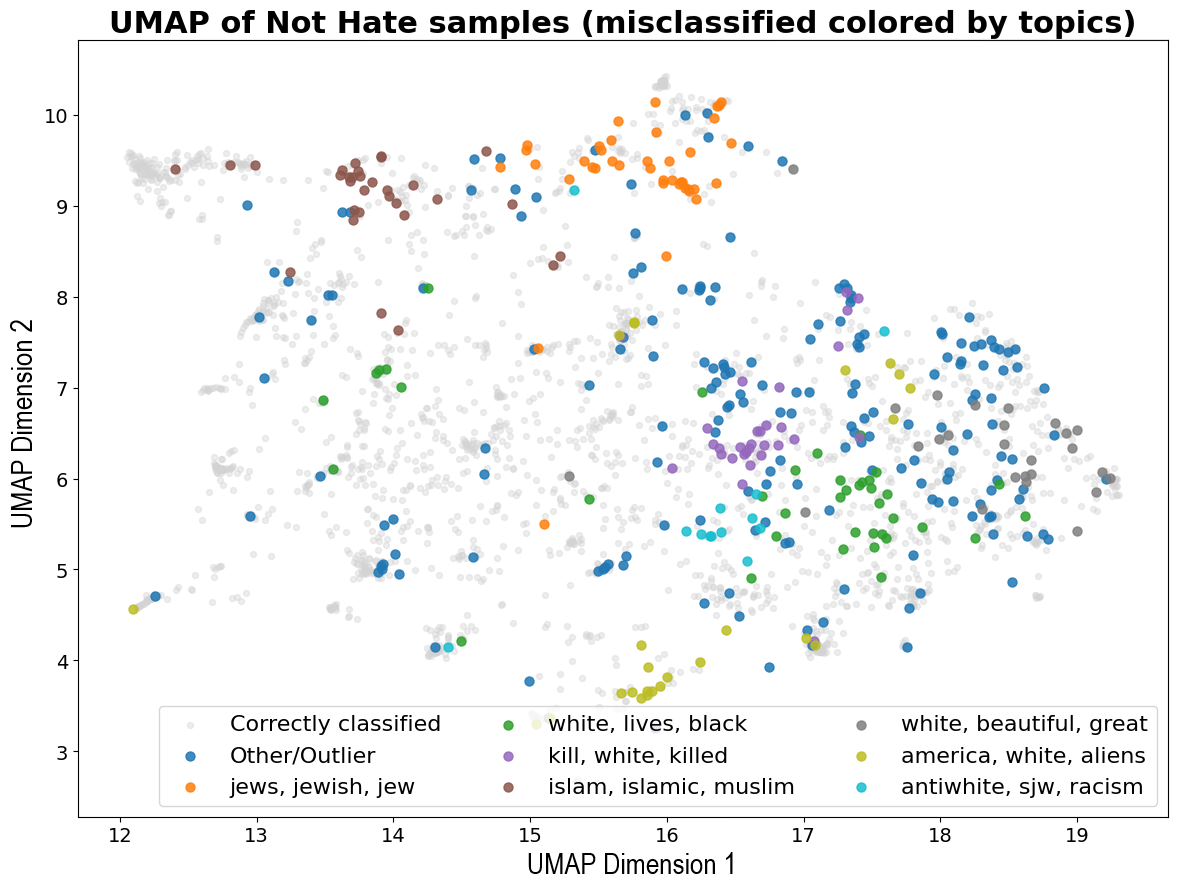}

    \caption{Predictions from NV-Embed-based classifier for the \texttt{hate} and \texttt{not hate} classes of IHC~\cite{elsherief-etal-2021-latent}. Different colors indicate different misclassified topics, and gray indicates correctly classified samples. UMAP~\cite{mcinnes2018umap} was used to project the embeddings into a 2D space for visualization.}
    \Description{Two side-by-side scatter plots showing the topics of misclassified samples.}
    \label{fig:cluster_topics1}
\end{figure*}

The main misclassified topics are shown in Figure~\ref{fig:topic_diagram}, which also displays the proportions of misclassification types. This allows us to understand which topics tend to be wrongly classified in a specific category. 
Figure~\ref{fig:cluster_topics1} provides a visualization of the sample distribution within the topics and the overall relation between the topics discovered with BERTopic~\cite{grootendorst2022bertopic}.

\section{Limitations} 

Using embedding models demonstrate better generalization across different hate speech datasets, particularly in cross-dataset settings. 
We achieve state-of-the-art performance, but we hypothesize that more specialized components could further enhance its performance. GPU resources are also a concern for the application of our method. However, NV-Embed can be fine-tuned with a batch size of 1 using LoRA on a GPU with around 12 GB of memory, which is typical for consumer-grade GPUs. Furthermore, at inference time, it can be loaded using lower floating-point precision.

More extensive investigation is necessary before deploying embedding models in real-world moderation systems. While this work focuses on IHS detection, further research is needed to extend this approach to other moderation tasks in order to validate its broader generalization.

We acknowledge the ethical complexities of working with offensive content. However, it is crucial to proactively address online hate.
While we provide an initial analysis to explore the limitations of the proposed method, a more thorough investigation is needed, particularly given that the models are trained on datasets that are inherently biased. Current deep learning models are still too complex to be fully audited and lack sufficient security and reliability, and we emphasize the importance of responsible deployment.

\clearpage
\end{document}